# Reconstructing Multi-echo Magnetic Resonance Images via Structured Deep Dictionary Learning

*Vanika Singhal and Angshul Majumdar*

Indraprastha Institute of Information Technology

{vanikas and angshul}@iiitd.ac.in

**Abstract** – Multi-echo magnetic resonance (MR) images are acquired by changing the echo times (for T2 weighted) or relaxation times (for T1 weighted) of scans. The resulting (multi-echo) images are usually used for quantitative MR imaging. Acquiring MR images is a slow process and acquiring multi scans of the same cross section for multi-echo imaging is even slower. In order to accelerate the scan, compressed sensing (CS) based techniques have been advocating partial K-space (Fourier domain) scans; the resulting images are reconstructed via structured CS algorithms. In recent times, it has been shown that instead of using off-the-shelf CS, better results can be obtained by adaptive reconstruction algorithms based on structured dictionary learning. In this work, we show that the reconstruction results can be further improved by using structured deep dictionaries. Experimental results on real datasets show that by using our proposed technique the scan-time can be cut by half compared to the state-of-the-art.

## 1. Introduction

The problem of magnetic resonance (MR) image reconstruction is to recover the image from the K-space (Fourier frequency) scan acquired by the scanner. When the K-space is fully sampled the reconstruction is trivial; application of the inverse fast Fourier transform (FFT) solves the problem. However, acquiring the full K-space is time consuming. This affects both the patient undergoing the scan as well as the MR facility at the hospital.

For the patient, staying in the claustrophobic scanner for long is an ordeal. The incessant switching of the gradient coils creates the condition even worse. Therefore, from the patient's point of view, a faster scan (like that in X-Ray computed tomography – lasting a few minutes) would be more welcome. For the hospital, slower acquisition means fewer patients to scan which in turn means lower revenue. For a purely commercial purpose, hospitals and other healthcare centers providing MR facility would like to speed-up the scans.

In order to ameliorate the situation reducing the scan-time of MR scanners has been a matter of utmost importance since the first commercial scanners became operational. Traditionally (before mid 2000's) changes in hardware were made in order to speed-up scans. However, they reached their limits soon. Since the later part of the 2000's software based acceleration techniques have proven to be promising. This has been largely fueled by the subject of compressed sensing (CS) [1, 2]. Since the first paper on this topic [3], a large number of papers have been published on this topic; a comprehensive (but slightly dated) treatise on the subject is available in [4].

A standard (single echo) brain scan takes about 20 minutes approximately. Imagine the time taken by a multi-echo acquisition protocol where 16 or 32 scans will be taken for each slice of the brain. This is the reason, reducing the scan time for multi-echo imaging is of utmost imaging. Yet there are only a handful of studies on this topic (except [5]), and all of them are covered in the book [3]; the major techniques will be discussed later.

CS basically exploits the transform domain sparsity of the image in order to reconstruct it from partial K-space scans. In doing so, it relied on defined transforms like wavelet or curvelet or DCT. However, later studies showed that instead of using fixed transforms, if the sparsity basis can be learnt from the data, better reconstruction results can be achieved. Such techniques are broadly dubbed as dictionary learning [6-8]. Such dictionary learning techniques were also adopted for solving multi-echo MR reconstruction [5].

Fueled by the success of deep learning, a very recent work [9] showed that instead of learning a single layer of dictionary, much better results can be obtained when multiple deep layers of dictionaries are learnt. The said work applied it for the single echo reconstruction problem. In this work, we will show that by modifying [9] in order to account for structure in multi-echo MR images, much better results can be obtained as compared to the state-of-the-art.

## 2. Literature Review

### 2.1. Single echo MR image reconstruction

Before going into the details of multi-echo reconstruction, we will briefly discuss the prevalent techniques in single echo MR image reconstruction. The discussion in this sub-section is elaborated in [4]. The image acquisition for partial K-space sampling is expressed as follows –

$$y = RFx + \eta \tag{1}$$

Here x is the image being acquired (and need to be reconstructed), F is the Fourier transform, R is the partial sampling mask and y the acquired K-space samples. The process is usually corrupted by Gaussian noise denoted by η.

Reconstructing x from (1) turns out to be an under-determined linear inverse problem which has infinitely many solutions. In CS the sparsity of x in some transform domain (wavelet, DTC etc.) is exploited in order to reconstruct it. The optimization problem is posed as,

$$\min_{\alpha} \|y - RF\Psi\alpha\|_2^2 + \lambda \|\alpha\|_1 \tag{2}$$

Here Ψ is the sparsifying domain and α the sparse coefficients. Once the $l_1$-minimization problem is solved, the image is obtained by inverting the sparse coefficients.

The $l_1$-minimization problem (2) is called the sparse synthesis prior formulation; there is an alternate co-sparse analysis prior formulation. In the analysis prior, the image is directly reconstructed such that the image remains co-sparse in some transform domain. The optimization problem turns out to be,

$$\min_{x} \|y - RF\alpha\|_2^2 + \lambda \|\Psi^T x\|_1 \tag{3}$$

It can be easily shown that for orthogonal transforms, the two formulations (2) and (3) will yield the same result, but for tight-frames the two formulations will yield different results. The co-sparse analysis formulation is more generic since it can accommodate for any linear transform, not necessarily orthogonal or tight-frame; this allows for well known image priors like total variation.

In standard CS, the sparsifying transform / basis $\Psi$ is fixed. It is well known in signal processing that compared to such fixed basis, better results can be obtained by learning the basis adaptively from the data. This led to the application of dictionary learning techniques in reconstruction. This is expressed as follows –

$$\min_{x,D,Z} \underbrace{\|y - RFx\|_2^2}_{\text{data fidelity}} + \lambda \underbrace{\left( \sum_i \|P_i x - D z_i\|_2^2 + \gamma \|z_i\|_1 \right)}_{\text{dictionary learning}} \quad (4)$$

The data fidelity term remains the same as before. But the sparsifying basis D is learnt from patches of the image $P_i x$ such that the learnt representation $z_i$ is sparse.

This (4) is the synthesis formulation, since the dictionary is learnt such that it synthesizes the samples from the learnt coefficients. There is an alternate analysis formulation where the basis operates on the data to produce the coefficients. This is called transform learning. The reconstruction formulation is expressed as,

$$\min_{x,T,Z} \underbrace{\|y - RFx\|_2^2}_{\text{data fidelity}} + \lambda \underbrace{\left( \sum_i \|T P_i x - z_i\|_2^2 + \mu \left( \|T\|_F^2 - \log \det(T) \right) + \gamma \|z_i\|_1 \right)}_{\text{transform learning}} \quad (5)$$

The data fidelity term does not change. The term within the brackets represent transform learning. The transform T operates on patches of the image $P_i x$ in order to produce sparse coefficients $z_i$. The regularization on T is required to prevent degenerate and trivial solutions.

This concludes the standard approaches in single echo MTI reconstruction. There are variants of such CS based techniques such as [10, 11]; but it is not possible to discuss all such variants.

## 2.2. Multi-echo MR image reconstruction

In multi-echo MR imaging, the same cross section is acquired by varying either T1 weighting or T2 weighting. This is achieved by varying the relaxation time (for T1) or echo time (for T2). For details please peruse [4]. The image acquisition can be expressed as follows –

$$y_i = R_i F x_i + \eta_i, \quad i = 1...n \quad (6)$$

In a combined fashion, this can be represented as,

$$\begin{bmatrix} y_1 \\ ... \\ y_n \end{bmatrix} = \begin{bmatrix} R_1 F & ... & 0 \\ ... & ... & ... \\ 0 & ... & R_n F \end{bmatrix} \begin{bmatrix} x_1 \\ ... \\ x_n \end{bmatrix} + \begin{bmatrix} \eta_1 \\ ... \\ \eta_n \end{bmatrix} \quad (7)$$

It turns out that the multi-echo images have a common co-sparse support which leads to a group-sparse structure in the transform domain. Incorporating the sparsifying transform into (7) leads to,

$$\begin{bmatrix} y_1 \\ \ldots \\ y_n \end{bmatrix} = \begin{bmatrix} R_1 F \Psi & \ldots & 0 \\ \ldots & \ldots & \ldots \\ 0 & \ldots & R_n F \Psi \end{bmatrix} \begin{bmatrix} \alpha_1 \\ \ldots \\ \alpha_n \end{bmatrix} + \begin{bmatrix} \eta_1 \\ \ldots \\ \eta_n \end{bmatrix} \quad (8)$$

The images x1 to xn are correlated, since they are from the same cross section and hence are structurally the same. Sparsifying transforms capture the structure / edges in the image in terms of high values along the structures and zeroes elsewhere. Therefore the sparse coefficients $\alpha_1, \ldots, \alpha_n$ has a common support, i.e. they will have high non-zero values at the same positions and zeroes elsewhere. Therefore when grouped by their indices, they will only have a few groups which are non-zero while the rest of the groups will be zeroes. Such a group-sparse solution is obtained by $l_{2,1}$-minimization [12].

$$\min_{\alpha_1,..\alpha_n} \left\| \begin{bmatrix} y_1 \\ \ldots \\ y_n \end{bmatrix} - \begin{bmatrix} R_1 F \Psi & \ldots & 0 \\ \ldots & \ldots & \ldots \\ 0 & \ldots & R_n F \Psi \end{bmatrix} \begin{bmatrix} \alpha_1 \\ \ldots \\ \alpha_n \end{bmatrix} \right\|_2^2 + \lambda \left\| \begin{bmatrix} \alpha_1 \\ \ldots \\ \alpha_n \end{bmatrix} \right\|_{2,1} \quad (9)$$

Alternately, one can also have an analysis prior formulation –

$$\min_{x_1,...x_n} \left\| \begin{bmatrix} y_1 \\ \ldots \\ y_n \end{bmatrix} - \begin{bmatrix} R_1 F & \ldots & 0 \\ \ldots & \ldots & \ldots \\ 0 & \ldots & R_n F \end{bmatrix} \begin{bmatrix} \Psi^T x_1 \\ \ldots \\ \Psi^T x_n \end{bmatrix} \right\|_2^2 + \lambda \left\| \begin{bmatrix} \Psi^T x_1 \\ \ldots \\ \Psi^T x_n \end{bmatrix} \right\|_{2,1} \quad (10)$$

Discussion on variants of these techniques is available in [4]. A later study on structured adaptive basis learning [5], showed how one can improve the results by adaptively learning the basis. The synthesis formulation was designed such that the learnt coefficients from each echo have a common sparse support. This ultimately led to group-sparse dictionary learning.

$$\min_{X,D,Z_i} \left\| vec(Y) - \begin{bmatrix} R_1 F & \ldots & 0 \\ \ldots & \ldots & \ldots \\ 0 & \ldots & R_n F \end{bmatrix} vec(X) \right\|_2^2 + \lambda \left( \sum_i \| P_i X - D Z_i \|_F^2 + \gamma \| Z_i \|_{2,1} \right) \quad (11)$$

where $vec(Y) = \begin{bmatrix} y_1 \\ \ldots \\ y_n \end{bmatrix}$, $vec(X) = \begin{bmatrix} x_1 \\ \ldots \\ x_n \end{bmatrix}$, $X = [x_1 | \ldots | x_1]$, $X_i = P_i X$ and $Z_i = [z_1 | \ldots | z_n]$.

An alternate transform learning based formulation was proposed in [2] as well. This was expressed as,

$$\min_{X,T,Z_i} \left\| vec(Y) - \begin{bmatrix} R_1 F & \ldots & 0 \\ \ldots & \ldots & \ldots \\ 0 & \ldots & R_n F \end{bmatrix} vec(X) \right\|_2^2 + \lambda \left( \sum_i \| T P_i X - D Z_i \|_F^2 + \mu \left( \| T \|_F^2 - \log \det(T) \right) + \gamma \| Z_i \|_{2,1} \right) \quad (12)$$

The symbols have already been defined.

To the best of our knowledge this [5] is the latest work on multi-echo MRI reconstruction; it is known to yield the best possible results.

### 2.3. Deep dictionary learning reconstruction

In (shallow) dictionary learning, the patches of the image are expressed as a product of one dictionary and corresponding coefficients. In deep dictionary learning, instead of one layer of dictionary there are multiple layers. The basic formulation for MR image reconstruction based on this concept has been published recently [6]. The optimization problem is expressed as,

$$\min_{x,D_1,D_2,D_3,z_i} \|y - RFx\|_2^2 + \lambda \left( \underbrace{\sum_i \|P_i x - D_1 \varphi(D_2 \varphi(D_3 z_i))\|_2^2 + \gamma \|z_i\|_1}_{\text{deep dictionary learning}} \right) \quad (13)$$

The formulation (13) is shown for three levels of dictionaries. There is a non-linearity between each layer in order to prevent the dictionaries from collapsing into a single layer. With deep dictionary learning, the results improve compared to single layer of dictionary.

## 3. Proposed Structured Deep Dictionary Learning

In this work we propose two variants. The first one is the extension of joint-sparse dictionary learning to the deep dictionary learning framework. The second one is to formulate a low-rank version of deep dictionary learning in order to preserve the similarities between multiple echoes.

### 3.1. Row-sparse Deep Dictionary Learning

In this version, we simply extend upon [5]. We learn multiple layers of dictionaries so that the learnt coefficients from different echoes will have a common sparse support. For understanding the reason behind this argument, we request the reader to peruse [5]; explaining the reason behind this assumption is beyond the scope of this work.

$$P_i[x_1 | ... | x_n] = P_i X_i = D_1 D_2 D_3 Z_i \quad (14)$$

Here we are representing the patches from different echoes stacked vertically as a product of multiple layers of dictionaries and corresponding coefficients. The coefficient matrix Z is formed by stacking the patch-wise coefficients as columns. In order to preserve the structure of multi-echo MRI, we will impose row-structure on Z. Note that we do not explicitly show the non-linearity between the layers; we will use ReLU (rectified linear unit) type non-linearity implicitly during the learning process.

The corresponding optimization problem will turn out to be,

$$\min_{X,D_1,D_2,D_3,Z_i} \left\| vec(Y) - \begin{bmatrix} R_1 F & ... & 0 \\ ... & ... & ... \\ 0 & ... & R_n F \end{bmatrix} vec(X) \right\|_2^2 + \lambda \left( \sum_i \|P_i X - D_1 D_2 D_3 Z_i\|_F^2 + \gamma \|Z_i\|_{2,1} \right) \quad (15)$$

s.t. $D_2 D_3 Z_i > 0$ and $D_3 Z_i > 0$

Here $P_iX$ is the patch from X and Z is the combined coefficient matrix from all patches. The constraints impose ReLU type non-linearity.

The complete problem (15) will be solved using the variable splitting approach [13]. We introduce two proxy variables

$D_2 D_3 Z_i = Z_i^1$ and $D_3 Z = Z_i^2$

With these proxies the corresponding augmented Lagrangian takes the form –

$$\min_{X, D_1, D_2, D_3, Z_i, Z_i^1, Z_i^2} \left\| vec(Y) - \begin{bmatrix} R_1 F & \cdots & 0 \\ \cdots & \cdots & \cdots \\ 0 & \cdots & R_n F \end{bmatrix} vec(X) \right\|_2^2$$
$$+ \lambda \left( \sum_i \|P_i X - D_1 Z_i^1\|_F^2 + \gamma \|Z_i\|_{2,1} + \mu_1 \|Z_i^1 - D_2 Z_i^2\|_F^2 + \mu_1 \|Z_i^2 - D_3 Z_i\|_F^2 \right) \quad (16)$$
$s.t.\ Z_i^1 > 0$ and $Z_i^2 > 0$

Using alternating direction method of multipliers (ADMM) [14], (16) can be segregated into the following sub-problems.

$$P1: \min_X \left\| vec(Y) - \begin{bmatrix} R_1 F & \cdots & 0 \\ \cdots & \cdots & \cdots \\ 0 & \cdots & R_n F \end{bmatrix} vec(X) \right\|_2^2 + \lambda \left( \sum_i \|P_i X - D_1 Z_i^1\|_F^2 \right)$$

$$P2: \min_{D_1} \left( \sum_i \|P_i X - D_1 Z_i^1\|_F^2 \right)$$

$$P3: \min_{D_2} \left( \sum_i \|Z_i^1 - D_2 Z_i^2\|_F^2 \right)$$

$$P4: \min_{D_3} \left( \sum_i \|Z_i^2 - D_3 Z_i\|_F^2 \right)$$

$$P5: \min_{Z^1} \left( \sum_i \|P_i X - D_1 Z_i^1\|_F^2 + \mu_1 \|Z_i^1 - D_2 Z_i^2\|_F^2 \right) s.t.\ Z^1 > 0$$

$$P6: \min_{Z^2} \left( \sum_i \mu_1 \|Z_i^1 - D_2 Z_i^2\|_F^2 + \mu_1 \|Z_i^2 - D_3 Z_i\|_F^2 \right) s.t.\ Z^2 > 0$$

$$P7: \min_Z \left( \sum_i \|P_i X - D_1 Z_i^1\|_F^2 + \gamma \|Z_i\|_{2,1} \right)$$

All the sub-problems from P1 to P6 are linear least square problems having a closed form solution. In sub-problems P5 and P6 the constraints are simply satisfied by putting the negative values of $Z^1$ and $Z^2$ to zeroes after obtaining the analytic solution. Sub-problem P7 is a row-sparse multiple measurement vector recovery problem that can be solve efficiently using algorithms in [15].

The problems are non-convex hence there is no guarantee of convergence to a global minimum. But recent works like [14] have showed that ADMM converges to a local minima for non-convex non-smooth (owing to the $l_{2,1}$-norm) problems such as ours.

## 3.2. Low-rank Deep Dictionary Learning

In the previous formulation we have considered that the $Z_i$ formed by stacking the coefficients of the $i^{th}$ patch of each echo will for a row-sparse matrix. This formulation stems from the assumption that the coefficients will have a common sparse support. There can be an alternate assumption, where we postulate that $Z_i$ will be a low-rank matrix. Such an assumption, although not very common for multi-echo MR reconstruction has been used for a similar problem arising in multi-channel parallel MRI [16, 17]. Both multi-echo MRI and multi-channel MRI are examples of multiple measurement vector recovery problems where the vectors look similar to each other. In group-sparsity the similarity is exploited by joint-sparsity while in low-rank techniques, the similarity is assumed to lead to linearly dependent columns of a matrix leading to rank deficiency.

With the said assumption, the only difference in formulation comes in the regularization of $Z_i$. Instead of the $l_{2,1}$-norm as used before (15), we need to employ the nuclear norm.

$$\min_{X,D_1,D_2,D_3,Z_i} \left\| vec(Y) - \begin{bmatrix} R_1 F & \ldots & 0 \\ \ldots & \ldots & \ldots \\ 0 & \ldots & R_n F \end{bmatrix} vec(X) \right\|_2^2 + \lambda \left( \sum_i \| P_i X - D_1 D_2 D_3 Z_i \|_F^2 + \gamma \| Z_i \|_* \right) \quad (17)$$

s.t. $D_2 D_3 Z_i > 0$ and $D_3 Z_i > 0$

The approach to solve (17) remains exactly the same as before; therefore we do not repeat it. We employ the same proxies and split the augmented Lagrangian in the same fashion as before using ADMM. The sub-problems P1 to P6 remain the same. Only the sub-problem P7 changes to the following,

$$\min_{Z_i} \| P_i X - D_1 Z_i^1 \|_F^2 + \gamma \| Z_i \|_* \quad (18)$$

This is a standard nuclear norm minimization problem that can be efficiently solved using singular value shrinkage [18].

For both the problems variable splitting introduces two hyper-parameters $\mu_1$ and $\mu_2$. In general they are supposed to be tuned. However, in this particular context they carry a special meaning; they are the relative importance weights for first and second layer of dictionary coefficients. Since there is no reason to prefer one layer over the other, we argue that these hyper-parameters can be set to unity.

## 4. Experimental Results

### 4.1. Data and Processing

All animal experimental procedures were carried out in compliance with the guidelines of the Canadian Council for Animal Care and were approved by the institutional Animal Care Committee. One female Sprague-Dawley rat was obtained from a breeding facility at the University of British Columbia and acclimatized for seven days prior to the beginning of the study. Animal was deeply anaesthetized and perfused intracardially with phosphate buffered saline for 3 minutes followed by freshly hydrolysed paraformaldehyde (4%) in 0.1 M sodium phosphate buffer at pH 7.4. The 20 mm spinal cord centred at C5 level was then harvested and post-fixed in the same fixative. MRI experiments were carried out on a 7 T/30 cm bore animal MRI scanner (Bruker, Germany). Single slice multi-echo CPMG (Carr-Purcell-Meiboom-Gill) sequence (8) was used to acquire fully sampled *k*-space data from the excised spinal cord sample using a 5 turn, 13 mm inner diameter solenoid coil with 256 × 256 matrix size, TE/TR = 6.738/1500 ms, 32 echoes, 2.56 cm field-of-view (FOV), 1 mm slice, number of averages (NA) = 8, and the excitation pulse phase cycled between 0° and 180°. Acquisition time was 50 minutes. The high signal average (NA = 8) was used to minimize noise contribution to better assess the efficacy of our CS scheme in the simulated undersampling portion.

Rectangular coil 22 x 19 mm was surgically implanted over the lumbar spine (T13/L1) of a female Sprague-Dawley rat as described previously. For MRI experiments, animal was anaesthetized with isofluorine (5% induction, 2% maintenance) mixed with medical air and positioned supine in a specially designed holder. Respiratory rate and body temperature were monitored using an MRI compatible monitoring system (SA Instruments, Stony Brook, NY). Heated circulating water was used to maintain the body temperature at 37°C. Data was acquired using the same CPMG sequence but with slice thickness of 1.5 mm and in-plane resolution of 117 μm. The slice was positioned at T13/L1 level, and NA=6. To minimize motion artefacts the acquisition was triggered to the respiratory rate, which resulted in the total acquisition time of approximately 45 minutes.

Undersampled K-space data in the phase encode direction was generated for each set of data for different acceleration factors (4 and 8, which corresponds to 64 and 32 phase encoding lines) from the fully sampled K-space. 33% of the read-out lines were placed around the centre of the K-space, and the rest randomly distributed in the periphery up to the desired number of phase encoding steps for the prescribed acceleration factor. Different sampling patterns were used for each echo.

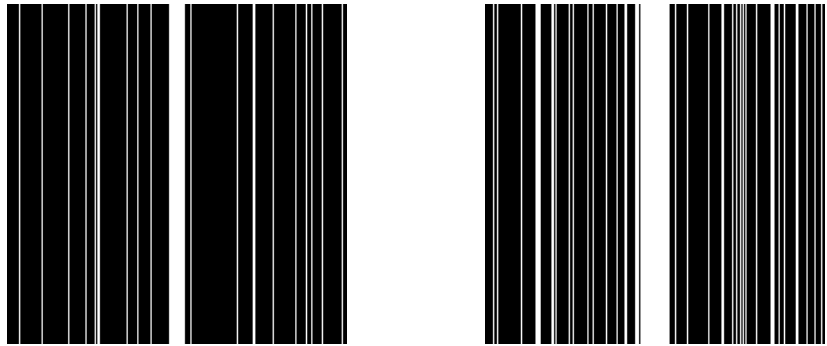

Fig. 1. Examples Sampling Masks. Left – 32 Lines. Right – 64 lines.

4.2.   Quantitative results

The prior work [2] already established that it yields better results than CS techniques. Therefore, we will compare only with the techniques in [2]. As discussed in section 2.2. there are two variants based on shallow dictionary learning (DL) and shallow transform learning (TL).

We have proposed two techniques here – row-sparse deep dictionary learning (RSDDL) and low-rank deep dictionary learning (LRDDL). Both the techniques requires specification of several parameters. We have used the greedy L-curve technique [19, 20] to tune these.

The comparative results are shown tables 1 and 2. We have shown the results for 2, 3 and 4 layers. With two layers or four layers, the results are worse than three layers. We use overlapping patches of size 12x12 and the number of dictionary atoms are halved in each layer.

The comparison is done in terms of signal-to-noise ratio (SNR) computed between the reconstructed and the ground-truth (fully sampled) images. The results show that our method is far superior. What is interesting to note is that the results obtained by the previous technique with 64 lines is the same or worse compared to the results obtained from our method with 32 lines. This means that potentially, our technique can speed up multi-echo MRI scans by a factor of 2.

Table 1. Ex-Vivo: SNR from different techniques

| Recovery Method | 32 lines | 64 lines |
|---|---|---|
| Proposed RSDDL (2 layer) | 21.5 | 23.9 |
| Proposed LRDDL (2 layer) | 21.2 | 23.5 |
| Proposed RSDDL (3 layer) | **24.7** | 26.7 |
| Proposed LRDDL (3 layer) | **24.2** | 24.1 |
| Proposed RSDDL (4 layer) | 23.0 | 25.8 |
| Proposed LRDDL (4 layer) | 22.7 | 23.9 |
| Row-sparse DL formulation [2] | 19.2 | **23.6** |
| Row-sparse TL formulation [2] | 20.0 | **24.2** |

Table 4. In-Vivo: SNR from different techniques

| Recovery Method | 32 lines | 64 lines |
|---|---|---|
| Proposed RSDDL (2 layer) | 19.9 | 22.5 |
| Proposed LRDDL (2 layer) | 19.5 | 22.5 |
| Proposed RSDDL (3 layer) | **23.7** | 25.3 |
| Proposed LRDDL (3 layer) | **23.6** | 24.5 |
| Proposed RSDDL (4 layer) | 23.2 | 24.9 |
| Proposed LRDDL (4 layer) | 22.4 | 24.0 |
| Row-sparse DL formulation [2] | 16.5 | **22.3** |
| Row-sparse TL formulation [2] | 17.1 | **23.3** |

4.3. Qualitative Evaluation

In reconstruction experiments it is customary to show the reconstructed images for visual evaluation. However, it is difficult for non-experts to judge the quality of reconstruction only from visual images. Therefore, papers usually also show the difference (between ground-truth and reconstructed) images. From the difference image it is easier to evaluate the reconstruction quality; the darker the images the better is the reconstruction quality.

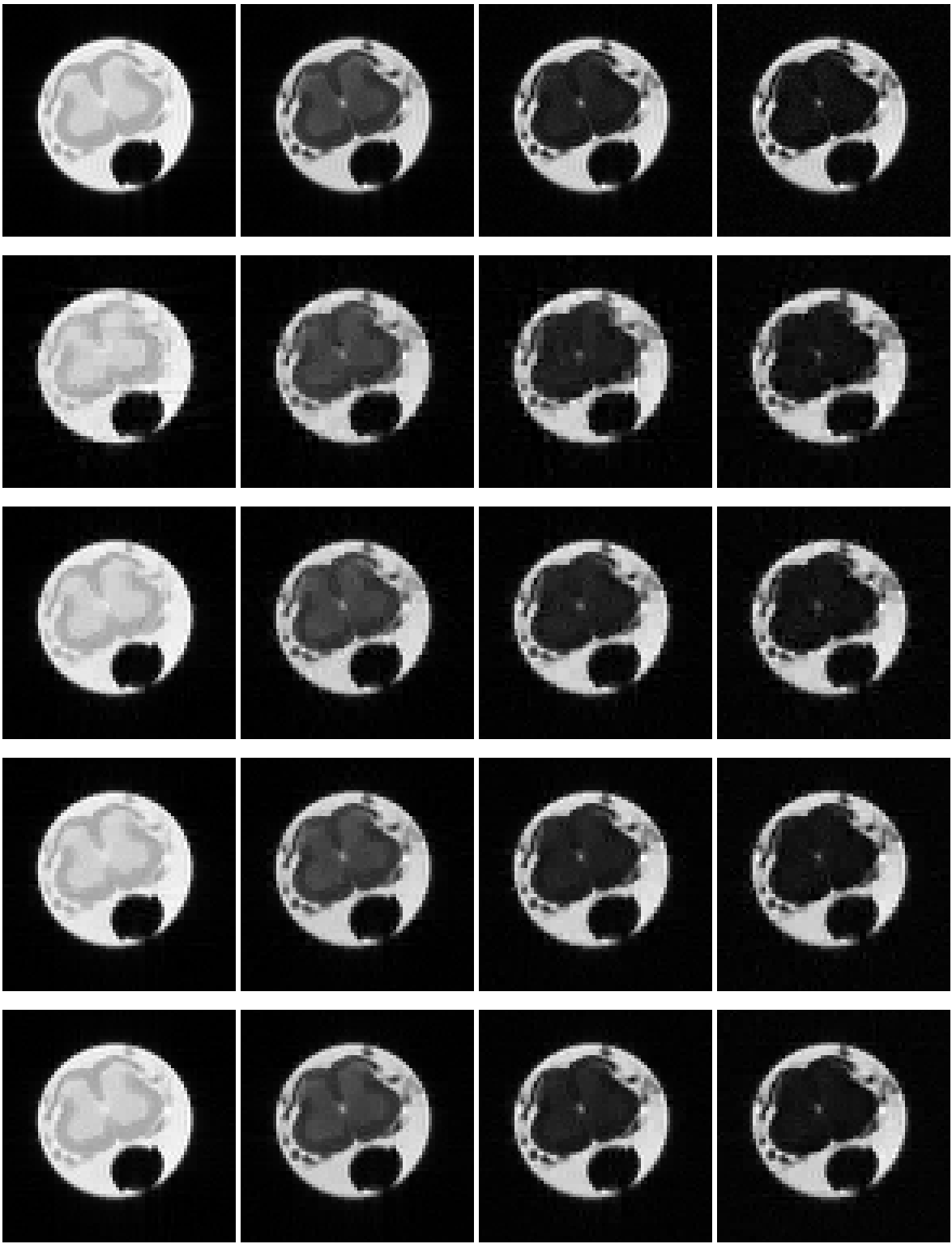

Fig. 3. Images from Ex-vivo Reconstruction. Top row –groundtruth; 2$^{nd}$ row – DL [2]; 3$^{rd}$ row – TL [2], 4$^{th}$ row – Proposed RSDDL (3 layers); 5$^{th}$ row – Proposed LRDDL (3 layers).

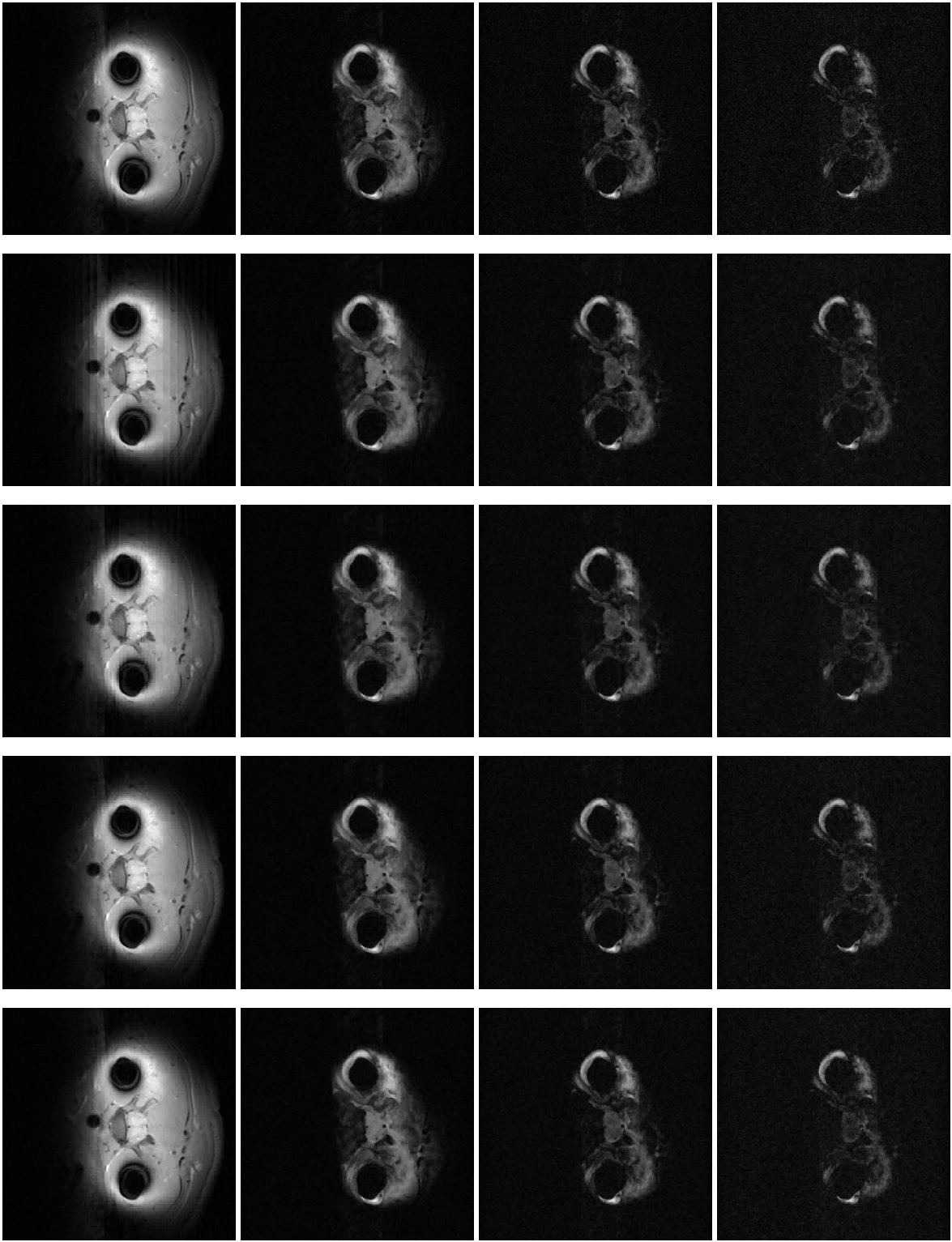

Fig. 4. Images from In-vivo Reconstruction. Top row –groundtruth; $2^{nd}$ row – DL [2]; $3^{rd}$ row – TL [2], $4^{th}$ row – Proposed RSDDL (3 layers); $5^{th}$ row – Proposed LRDDL (3 layers).

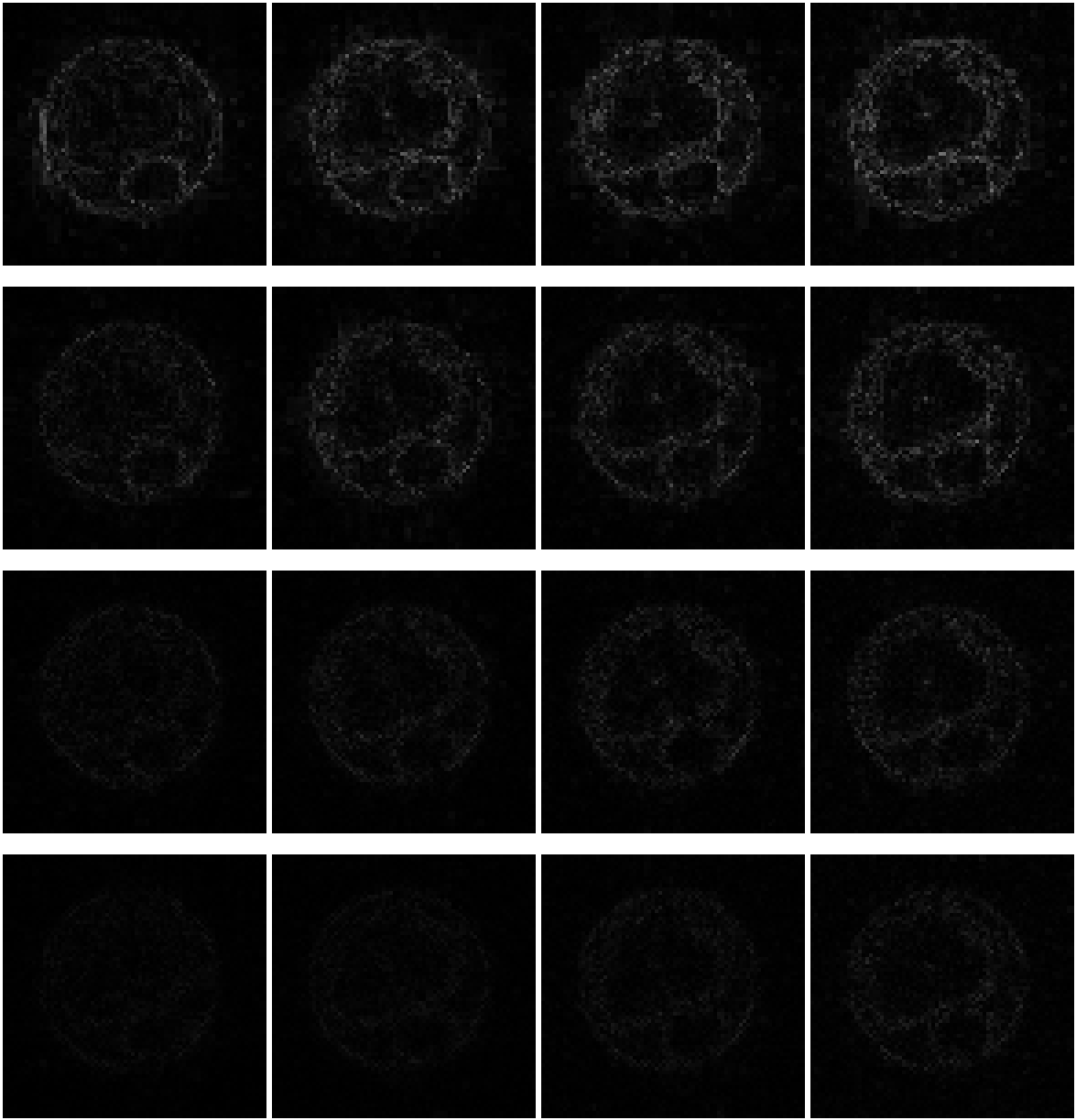

Fig. 5. Difference Images from Ex-vivo Reconstruction. Top row –groundtruth; 2$^{nd}$ row – DL [2]; 3$^{rd}$ row – TL [2], 4$^{th}$ row – Proposed RSDDL (3 layers); 5$^{th}$ row – Proposed LRDDL (3 layers).

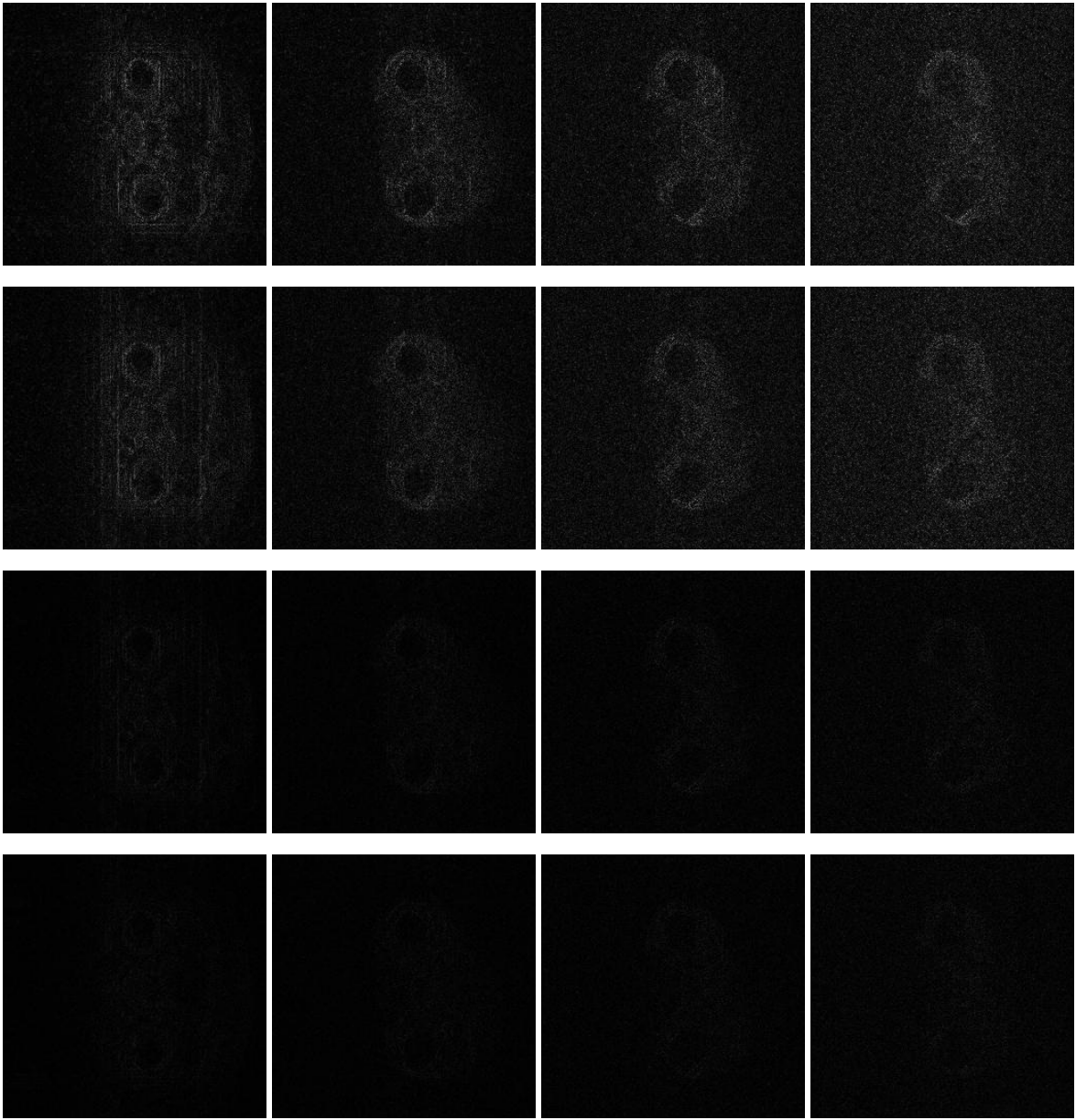

Fig. 6. Difference Images from In-vivo Reconstruction. Top row –Analysis row-sparsity [3]; 2$^{nd}$ row – Rank deficient analysis row-sparsity [24], 3$^{rd}$ row – Sparse DL formulation, 4$^{th}$ row – Proposed row-sparse DL

In Fig.s 2 and 3 we show the reconstructed images from 4 random echoes from ex-vivo and in-vivo datasets and in Fig.s 4 and 5 we show the difference images (for the same echoes). The results are shown for 3 layers since it yields the best reconstruction. These images correspond to reconstruction from 32 lines (acceleration factor 8).

From the reconstructed images we can see that with the shallow techniques from [2] there are heavy reconstruction artifacts appearing as vertical stripes. This is especially discernible in the brighter portions of the images. From our proposed method the artifacts are almost indiscernible. The superiority in reconstruction quality is corroborated in the difference images. They are more whitish for the techniques from [2] but are almost completely dark from our proposed methods. This goes to show the superior reconstruction quality of our proposed deep techniques.

## 5. Conclusion

The objective of this work is to improve reconstruction of multi-echo MRI images from partial K-space scans. Since the multi-echo images are correlated with each other, structured CS reconstruction techniques were initially proposed for reconstruction. More recent work showed that, instead of using CS directly, better results can be obtained by adaptively learning the basis. In this work, we show that instead of learning one level of basis, further improvement can be achieved when multiple layers of basis are learnt.

We proposed two variants. The first one is based on exploiting the inter-echo correlation as common sparse support and the second one is based on modelling the correlation as linear dependency leading to a low-rank penalty. Both the formulations yield similar results.

In this work, we proposed a synthesis formulation for structured deep dictionary learning reconstruction. This yields better results than shallow approaches. However, among the shallow approaches, it has been found that an analysis formulation yields better results [2]. We believe that extending the analysis formulation to a deeper version might improve the results even further.

As cursorily mentioned in the paper, there is another problem of MR imaging that is mathematically similar – multi-coil parallel MRI. In future, we would like to apply these techniques for parallel MRI reconstruction.